\documentclass[sigconf]{acmart}

\renewcommand\footnotetextcopyrightpermission[1]{} 
\usepackage{booktabs} 
\usepackage{amsmath}
\usepackage[linesnumbered,ruled]{algorithm2e}

\usepackage[subtle]{savetrees}

\setcopyright{acmcopyright}

\usepackage{arydshln}

\newcommand{\KILL}[1]{}
\newcommand{\HIDE}[1]{}

\setcopyright{none}

\begin{document}
\title{Semantic Relatedness for Keyword Disambiguation: \\ Exploiting Different Embeddings}

\author{Mar\'ia G. Buey}
\email{magrabue@doctor.upv.es}
\affiliation{%
  \institution{everis / NTT Data \\ Universitat Polit\`ecnica de Val\`encia}
  \city{Val\`encia}
  \country{Spain}
}
\author{Carlos Bobed}
\email{cbobed@unizar.es}
\affiliation{%
  \institution{everis / NTT Data \\ IIS Department, University of Zaragoza}
  \city{Zaragoza}
  \country{Spain}
}
\author{Jorge Gracia}
\email{jogracia@unizar.es}
\affiliation{%
  \institution{Aragon Institute of Engineering Research (I3A) / \\IIS Department, University of Zaragoza}
  \city{Zaragoza}
  \country{Spain}
}
\author{Eduardo Mena}
\email{emena@unizar.es}
\affiliation{%
  \institution{Aragon Institute of Engineering Research (I3A) / \\ IIS Department, University of Zaragoza}
  \city{Zaragoza}
  \country{Spain}
}

\begin{abstract}
Understanding the meaning of words is crucial for many tasks that involve human-machine interaction. This has been tackled by research in Word Sense Disambiguation (WSD) in the Natural Language Processing (NLP) field. Recently, WSD and many other NLP tasks have taken advantage of embeddings-based representation of words, sentences, and documents. However, when it comes to WSD, most embeddings models suffer from ambiguity as they do not capture the different possible meanings of the words. Even when they do, the list of possible meanings for a word (sense inventory) has to be known in advance at training time to be included in the embeddings space. Unfortunately, there are situations in which such a sense inventory is not known in advance (e.g., an ontology selected at run-time), or it evolves with time and its status diverges from the one at training time. This hampers the use of embeddings models for WSD. Furthermore, traditional WSD techniques do not perform well in situations in which the available linguistic information is very scarce, such as the case of keyword-based queries.

In this paper, we propose an approach to keyword disambiguation which grounds on a semantic relatedness between words and senses provided by an external inventory (ontology) that is not known at training time. Building on previous works, we present a semantic relatedness measure that uses word embeddings, and explore different disambiguation algorithms to also exploit both word and sentence representations. Experimental results show that this approach achieves results comparable with the state of the art when applied for WSD, without training for a particular domain.

\end{abstract}

%
%



\maketitle

\section{Introduction}
In any information system which requires user interaction, being able to understand the user is a crucial requirement, which is often tackled by limiting the user input (e.g., presenting predefined forms with fixed options). The more freedom you provide the user with, the more difficult interpretation the computer has to do to achieve a useful interaction. In such a context, being capable of disambiguating the input words (i.e., associating each word with its proper meaning in a given context) is the starting point of any interpretation process done by the computer. 

Usually, such a disambiguation process is tackled from a Natural Language Processing (NLP) perspective~\cite{Navigli2009}, assuming rich linguistic information, such as Part of Speech (POS), dependencies between words, etc.,  which is very useful to perform the task. However, due to the world wide use of Web search engines, users are very used to keyword interfaces and they still express their needs in such terms. In this scenario, although there are some studies which point out that \emph{keyword search queries} (aka., \emph{Web search queries}) exhibit their own language structure~\cite{barr2008linguistic,pinter2016syntactic,roy2016syntactic}, we still need methods to disambiguate the meanings of the words which do not need  such an information as it might not be available. 


Recent advances in NLP have focused on the development of different embedding models~\cite{bengio2003neural,mikolov2013distributed,le2014distributed,mancini2017embedding,camacho2016nasari,pennington2014glove}, which are a set of language modeling and feature learning techniques where elements from a vocabulary are mapped to a vectorial space capturing their \emph{distributional semantics}~\cite{sahlgren2008distributional}.
While there are different methods to build word embeddings, the latest (and most successful) word embedding techniques rely on neural network architectures~\cite{bengio2003neural,mikolov2013distributed,le2014distributed,pennington2014glove}. 
Their usage as the underlying input representation has boosted the performance of different NLP tasks~\cite{socher2013parsing,socher2013recursive}. However, in the context of disambiguation tasks, one of the main limitations of word embeddings is that the possible meanings of a word are combined into a single representation, i.e., a single vector in the semantic space. Such a limitation can be avoided by representing individual meanings of words as distinct vectors in the space (e.g., sense2vec~\cite{mancini2017embedding}). However, there are scenarios where we do not know all the different senses at training time (e.g., open domain scenarios where we cannot find all the possible meanings in a sense catalog), and, even if we would know them, we would require to have annotated data (which might be unavailable or expensive to obtain). Besides, we would need to train a model for each new scenario or new meaning that would be added to our catalog. Thus, we need a disambiguation method able to relate words and their senses in a flexible and general way (i.e., independently of the domain we are working in) by exploiting the available resources.

In this paper, we propose a keyword disambiguation method which is based on the semantic relatedness (the degree in which two objects are related by any kind of semantic relationship~\cite{budan2006evaluating}) between words, taking advantage of the semantic information captured by word embeddings. Our proposal makes possible to measure the relatedness not only among plain words but also among senses of words (which, in a Semantic Web context, can be expressed as ontological terms), and it is able to work independently of the resources used, i.e., the sense inventory whose meanings we want to map to and the word embedding model used as input.

For this purpose, we build on the work by Gracia~and~Mena on semantic relatedness~\cite{wise08} and disambiguation~\cite{gracia2009multiontology}. These works exploited the information about word co-occurrence frequencies provided by existing Web search engines. We evolve and adapt them to improve their performance using different kinds of embeddings (both at word and sentence level).
The main benefit of such an adaptation is two-fold:~1) we exploit the semantics captured by embeddings which goes further than co-occurrence of terms, and~2) we decouple the proposal from any Web search engine, being able to use off-the-shelf models trained by third parties for our purposes. This has an important side-effect: our measure can be easily adapted to any domain which we have a document corpus from. Indeed, this adaptation would require a training step, but it would be unsupervised and the only data required would be the corpus of documents itself. 

To evaluate our approach, we have carried out a thorough experimentation in the context of Word Sense Disambiguation (WSD), where we have used different pre-trained word embeddings publicly available on the Web, and WordNet\footnote{\label{wordnetfootnote}\url{https://wordnet.princeton.edu/}} as sense repository. Our measure improves the performance obtained in~\cite{wise08}, and achieves state of the art WSD values without the need of specific training for a specific sense inventory. This is especially relevant, for example, for systems based on keyword input and/or which have to work with dynamically selected ontologies~\cite{bobed2016querygen} or even with ontologies extracted directly from the Web~\cite{Halevy15}. All the experimental data and evaluation results are available online\footnote{\url{https://bit.ly/2lqCzop}}.

The rest of the paper is structured as follows. Section~\ref{sec:relatedwork} discusses related works. In Section~\ref{sec:approachdescription} we describe our semantic relatedness measure approximation, in  Section~\ref{sec:disambalgorithm} we present the disambiguation algorithm that we use, and Section~\ref{sec:experimentaleval} summarizes our experimental results. Finally, our conclusions and future work appear in Section~\ref{sec:conclusionsandfuturework}.

\section{Related Work}
\label{sec:relatedwork}
Semantic relatedness is the degree in which two objects are related by any kind of semantic relationship~\cite{budan2006evaluating} and lies at the core of many applications in NLP (such as WSD, Information Retrieval, Natural Language Understanding, or Entity Recognition). The term is often confused with semantic similarity, which measures the degree in which two objects are similar or equivalent. For example, "car" is similar to "bus", but is also related to "road" and "driving". It has received a great research interest and different types of methods have been developed: it can be statistically estimated (e.g. co-occurrence-based methods~\cite{landauer1998introduction}) and learned (e.g., distributional measures that estimate semantic relatedness between terms using a multidimensional space model to correlate words and textual contexts~\cite{mohammad2012distributional}); or it can be computed using a taxonomy or a graph (e.g., ontologies) to define the distance between terms or concepts~\cite{pirro2012reword}. Indeed, most methods rely on particular lexical resources (dictionaries, thesauri, or well structured taxonomies such as WordNet\textsuperscript{\ref{wordnetfootnote}}).

Regarding disambiguation, WSD methods can be classified into four conventional approaches: supervised~\cite{vial2018improving}, unsupervised~\cite{correa2018word}, semi-supervised~\cite{yuan2016semi}, and knowledge-based methods~\cite{chaplot2018knowledge}. For example, in a way similar to us, in the SemEval 2015 All-Words Sense Disambiguation and Entity Linking task\footnote{\label{semeval2015footnote}\url{http://alt.qcri.org/semeval2015/task13/}}, the majority of the approaches (LIMSI, SUDOKU, EBL-Hope, etc.) that best performed in WSD relied on the combination of unsupervised learning of semantic information from the content of a corpus (such as SemCor) and/or on lexical resources as sense inventories (such as WordNet or BabelNet) to disambiguate the sense of words in natural language sentences. However, to our knowledge, no previous works (excepting those of Gracia~and~Mena~\cite{wise08,gracia2009multiontology}) have studied specific methods for the disambiguation of words in keyword-based inputs, where the linguistic information is scarce.

Regarding the resources we use in our approach, word embeddings represent words in a low-dimensional continuous space and they are used to capture syntactic and semantic information from massive amounts of textual content. In recent years, they have gained great popularity due to this ability and many NLP applications have taken advantage of the potential of these distributional models. Bengio et al.~\cite{bengio2003neural} preceded a wide number of current language model techniques and several authors have proposed their own approaches~\cite{le2014distributed,pennington2014glove,camacho2016nasari} to construct word embeddings vectors where \emph{word2vec}~\cite{mikolov2013distributed} is the most widely used.

Despite their advantages, one of the main limitations of word embeddings is that possible meanings of a word are conflated into a single representation. Sense embeddings (e.g., sense2vec~\cite{mancini2017embedding}) are proposed as a solution to this problem: individual meanings of words are represented as distinct vectors in the space. These approaches are classified in two categories by how they model meaning and where they obtain it from~\cite{camacho2018word}: 1) unsupervised models which learn word senses from text corpora (by inducing different senses of a word, analyzing its contextual semantics in a text corpus and representing each sense based on the statistical knowledge derived from the corpus), and 2) knowledge-based methods which exploit sense inventories of lexical resources for representing meanings (such as WordNet\textsuperscript{\ref{wordnetfootnote}}, Wikipedia\footnote{\label{wikifootnote}\url{https://www.wikipedia.org/}}, BabelNet\footnote{\label{bnfootnote}\url{https://babelnet.org/}}, etc.). We can also find models that not only provide representations of words, but also the senses of the words in a joint embedded space. This is the case of NASARI vectors~\cite{camacho2016nasari} which  not only provide accurate representation of word senses in different languages, but they also include both concepts and named entities, all included in a single unified semantic space. 
However, in the first case (i.e., unsupervised models), we cannot target a particular sense inventory or ontology to perform the disambiguation, not having control for example about the concept detail/granularity, and, besides, the detected senses might not be aligned to any particular human-readable structure; in the second case, we need to know all the senses at training time, not being able to adapt to new scenarios (e.g., addition/deletion of senses in the directory, evolving ontologies, etc.). Thus, the sweet spot would be neither to require re-training nor newly labelled data, while being capable to perform the disambiguation against any sense repository.


Although sense embeddings capture and represent information about meanings and can be used to calculate the sense that a word has in a specific context, word embeddings have also been shown to have good performance in disambiguation tasks~\cite{iacobacci2016embeddings}. Therefore, we wanted to explore how we could push further the usage of word embeddings for keyword disambiguation. Working at word level (as starting point) allows us to use a semantic relatedness measure between terms and reuse available resources, without needing to train explicitly new word embeddings neither for a specific task nor for newly added possible senses (i.e., adapting to any given sense dictionary or ontology). We have taken as baseline the works presented in~\cite{wise08,gracia2009multiontology}. In~\cite{gracia2009multiontology}, the authors provide a keyword disambiguation algorithm that uses the semantic relatedness measure defined in~\cite{wise08} to find the appropriate sense for keywords. The authors focused on a method that exploits the Web as a source of knowledge, and a transformation of the Normalized Google Distance (NGD)~\cite{cilibrasi2007google} into a mixed way of relatedness measure (between ontology terms and plain words). 
We propose to, on the one hand, substitute this distance with a word embedding based one to take advantage of the semantics captured by embeddings, improving the performance regarding using just co-occurrence of terms; and, on the other hand, explore modifications of their algorithm to improve their disambiguation capabilities.

Finally, as pointed out by Lastra-D\'iaz~et~al.~\cite{lastra2019reproducible}, the embeddings that behave the best for disambiguation purposes are those which capture not only distributional semantics of texts, but also structural information about the possible meanings. We aim at achieving this disambiguation performance in a more flexible way, decoupling linguistic surface from the actual sense catalog (i.e., ontology) in order to adapt to new (i.e., unknown at training time) possible meanings, and being capable to apply it to keyword inputs, where the linguistic information is scarce.

\section{Relatedness Measure based on Word Embeddings}
\label{sec:approachdescription}
\begin{figure}
\includegraphics[width=8cm]{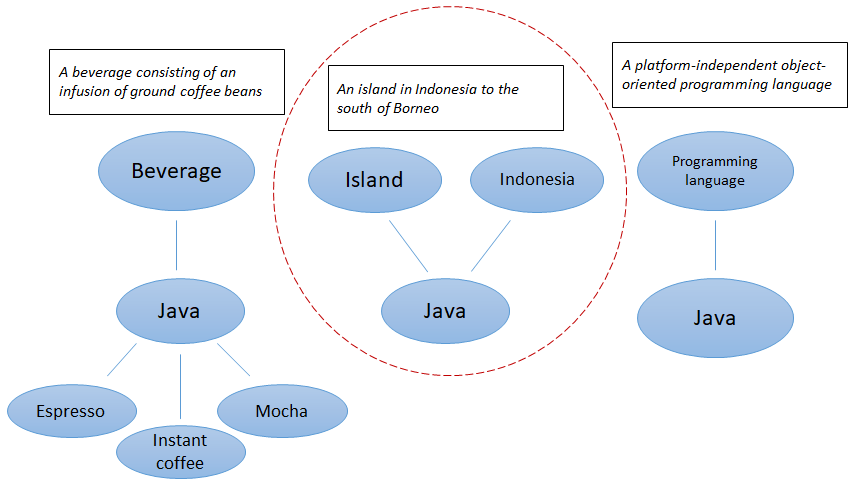}
\caption{Example of the semantic description of the term "Java" in WordNet.}
\label{fig:Java}
\end{figure}

Word embeddings can be used out-of-the-box to compute relatedness between words. However, they do not suffice in situations in which a relatedness has to be computed between senses (e.g., ontology terms in a Semantic Web context) or between senses and words. To that end, we ground on a previously defined relatedness measure between senses proposed by Gracia~and~Mena~\cite{wise08}. 
The authors proposed a method to compute the semantic relatedness between ontology terms (which we can see as individual senses), and an extension to calculate it between plain words and terms. 
Their proposal was built on the notion of the \emph{ontological context} of a term, which is constructed  combining the synonyms and the hypernyms of the ontological term (or sense). Given an ontological term ${t}$, they defined its \emph{ontological context} (denoted by ${OC(t)}$) as the minimum set of other ontological terms that belong to its semantic description, locating the term in the ontology and characterizing its meaning. For example, in the WordNet taxonomy, the class ``Java'' (in the sense of ``an Indonesian island''), is well characterized and distinguished from other senses by considering its direct hypernym ``Island'' (see Figure~\ref{fig:Java}).

Then, given two ontological terms $a$ and $b$, their relatedness measure is computed as: 

\begin{align}
\label{eq:completeeq}
\begin{split}
  rel(a,b)=w_0rel_0(a,b)+w_1rel_1(a,b), \\
  (w_0 \geq 0, w1 \geq 0, w_0+w_1=1) 
\end{split}
\end{align}

\noindent
with ${rel_0(a,b)}$ and ${rel_1(a,b)}$ computed as follows: 
\begin{align}
\label{eq:level0}
\begin{split}
rel_0(a,b)=\frac{\sum_{i, j}rel_{w}({syn_a}_i,{syn_b}_j)}{|Syn(a)||Syn(b)|}, \\
(i=1..|Syn(a)|, j=1..|Syn(b)|)
\end{split}
\end{align}
\begin{align}
\label{eq:level1}
\begin{split}
rel_1(a,b)=\frac{\sum_{i, j}rel_0({oc_a}_i,{oc_b}_j)}{|OC(a)||OC(b)|}, \\
(i=1..|OC(a)|, j=1..|OC(b)|)
\end{split} 
\end{align}

\noindent where ${rel_{w}}$ refers to the relatedness between words (as it will be defined later on in equations 7 and 9); ${Syn(a)=\{{syn_a}_1, {syn_a}_2, ...\}}$ and ${Syn(b)=\{{syn_b}_1, {syn_b}_2, ...\}}$ are the set of synonyms (equivalent labels, including the term label) of ontological terms \textit{a} and \textit{b}; ${OC(a)=}\\{\{{oc_a}_1, {oc_a}_2, ...\}}$ and ${OC(b)=\{{oc_b}_1, {oc_b}_2, ...\}}$ are the terms of their ontological context\footnote{Notice that ${|Syn(x)| \geq 1}$ and ${|OC(x)| \geq 0}$.}. Each \textit{a} and \textit{b} is characterized by taking into account two levels of their semantic description: \textit{Level 0)} the term label and its synonyms (Equation~\ref{eq:level0}), and \textit{Level 1)} its ontological context (Equation~\ref{eq:level1}). $w_0$ and $w_1$ are used to weight these levels\footnote{We set these values as $w_0 = w_1 = 0.5$ as indicated in~\cite{wise08}.}.

This measure can be also applied between an ontology term ${t}$ and a plain word ${w}$ which provides us with a value which indicates the relatedness degree between a sense and a word. So, in that case, the previous equations are computed as follows:
\begin{align}
\begin{split}
rel(t,w)=w_0rel_0(t,w)+w_1rel_1(t,w), \\ 
(w_0 \geq 0, w1 \geq 0, w_0+w_1=1)
\end{split} \label{eq:termWord}
\end{align}
\begin{equation}
rel_0(t,w)=\frac{\sum_{i,j}rel_{w}({syn_t}_i,w)}{|Syn(t)|}, \\
(i=1..|Syn(t)|)
\end{equation}
\begin{equation}
rel_1(t,w)=\frac{\sum_{i,j}rel_0({oc_t}_i,w)}{|OC(t)|}, \\
(i=1..|OC(t)|)
\end{equation}

Here, ${rel_{w}}$ is the distance that the authors used in~\cite{wise08} to measure how two plain words are related. They proposed a generalization of the Cilibrasi and Vit\'anyi's Normalized Google Distance \textit{NGD(x,y)}~\cite{cilibrasi2007google} to use any Web search engine as source of frequencies. This generalization is called Normalized Web Distance \textit{NWD(x,y)}, whose smaller values represent greater semantic relation between words. Although most of NWD values fall between 0 and 1, it ranges from 0 to ${\infty}$. Therefore, to obtain a proper relatedness measure in the range [0, 1] that increases inversely to distance, they proposed the following transformation:
\begin{equation}
\label{eq:relweb}
rel_w(x, y)=relWeb(x, y)=e^{-2NWD(x,y)}
\end{equation}

To explore the use of emerging word-embedding techniques in this context and compare them with those based on search engines, we propose to exploit the semantic capabilities of word embeddings in this formulation and substitute the ${relWeb(x,y)}$ measure. We could use the cosine similarity distance between the embedding vectors of the words, i.e., using the following equation:
\begin{equation}
sim(x,y)=cos(\theta)=\frac{X1 \bullet X2}{||X1||\cdot||X2||}
\end{equation}
\noindent where ${x}$ and ${y}$ are plain words, ${X1}$ and ${X2}$ their correspondent word embedding vectors, and ${\theta}$ the angle between them. However, ${sim(x,y)}$ ranges in [-1, 1], so, in order to obtain a 
distance in the range [0, 1] (so that we can substitute Equation~\ref{eq:relweb} directly in Equation~\ref{eq:level0}), we propose to use the \textit{angular distance} instead, which is computed as follows:
\begin{equation}
\label{eq:arccos}
rel_w(x, y)=ang. distance(x,y)=1-\frac{arccos(sim(x,y))}{\pi}
\end{equation}

\noindent So, in Equation~\ref{eq:level0}, we use Equation~\ref{eq:arccos} as ${rel_{w}}$ distance instead of Equation~\ref{eq:relweb}. 
We use this distance to compute the semantic relatedness between words, between ontology terms (or senses), or between ontology terms and words, obtaining a value between 0 and 1. For those cases in which the label of the ontological term is multi-word, we just compute the centroid of the set of words that form the label. While, at first, it might seem that we limit the coverage of the measure proposed in~\cite{wise08} (it built on the results of Web search engines, which potentially cover any domain), we have to bear in mind the plethora of word embedding models directly available in the Web, as well as the possibility of using our own corpus of documents to fine tuning our measure for a particular domain (which is easier to have, rather than crawling the whole WWW).

\section{Disambiguation Algorithm}
\label{sec:disambalgorithm}
We ground our keyword disambiguation proposal on the disambiguation algorithm defined in~\cite{gracia2009multiontology}, using the adapted semantic relatedness measure proposed in the previous section. This algorithm is based on the hypothesis that the most significant words in the disambiguation context are the most highly related to the word to disambiguate; such words conform the \emph{active context} of the word being disambiguated. 

As an overview, once the \emph{active context} of each input keyword has been calculated, the algorithm performs three main steps:~1)~obtaining the semantic relatedness between the active context of a keyword and its possible senses,~2) calculating the overlap between the words in the active context and the semantic descriptions (i.e., \emph{ontological context}) of the 
possible senses of the keyword to disambiguate, and 3) re-ranking the possible senses according to their frequency of use (only when such information is available for the sense inventory selected\footnote{If we do not have such information, we assume that all senses are equally likely.}).
Apart from using the updated $rel_w$ measure to select the \emph{active contexts}, we propose to modify the second step of this algorithm in order to study the influence of different approaches which exploit the semantic information captured by different word embeddings. 
In the following subsections, we first detail the original algorithm which we base our proposal on, and, then, we describe the modifications that we propose to improve its performance using word embeddings.

\subsection{Background: Algorithm Description}
\label{sec:algorithmdescription}
First of all, let us formally introduce the notion of \emph{active context}. Let~${k}$ be an element of an input sequence of words~${\mathbb{S}}$ with an intended meaning, ${\mathbb{K} \subseteq \mathbb{S}}$ be the set of all possible keywords in the input, ${C \subseteq \mathbb{K}}$ the set of keywords of the disambiguation context (i.e., the complete disambiguation window considered, e.g., the sentence where the keyword appear), and ${k_d \in \mathbb{K}}$ the target keyword to disambiguate. 
Thus:
\begin{definition}
Given a context $\textit{C} \in \mathbb{K}$, and a word to disambiguate ${k_d} \in \mathbb{K}$, the \textit{active context} ${C_a}$ of $k_d$ is a subset ${C_a \subseteq C}$ such that 
$\forall k_i \in C_a, \nexists k_j \in (C - C_a)~\ni rel(k_j,k_d) > rel(k_i,k_d)$.
\end{definition}
In other words, $C_a$ contains the words in the input that are the most related ones to $k_d$. 
To obtain such a context, we stick to the method proposed in~\cite{gracia2009multiontology}:~1)~removing repeated words and stopwords from \textit{C}, ~2) applying a semantic relatedness (${rel_{w}}$ in our case) between each context word ${k_i \in C}$ and the keyword to disambiguate ${k_d}$, and ~3) constructing ${C_a}$ with the context words whose relatedness scores above a certain threshold. The output of this process is the \textit{active context} ${C_a \subseteq C}$. The maximum cardinality of ${C_a}$ is set to a fixed  value(${|C_a| = 4}$) following Kaplan's experiments~\cite{kaplan1955experiment}. 

Once we have obtained $C_a$ for $k_d$, we 
can apply the main algorithm, which takes as input $k_d$, $C_a$, and a set of possible senses for $k_d$, ${S_{k_d}}$. 
The main steps are presented in Algorithm~\ref{alg:disambiguation}\footnote{We refer the interested reader to~\cite{gracia2009multiontology} for the complete details.}: 


\begin{algorithm}

\caption{Keyword disambiguation algorithm}\label{alg:disambiguation}
    \SetKwInOut{Input}{Input}
    \SetKwInOut{keyword}{$K_d$}
    \SetKwInOut{senses}{$S_{k_d}$}
    \SetKwInOut{context}{$C_a$}
    \SetKwInOut{Output}{Output}
    
    \Input{}
    \keyword{The keyword to disambiguate.}
    \senses{The set of possible senses for $K_d$.}
    \context{The active context selected for $K_d$.}
    \Output{A weight for each sense ${s_i \in S_{k_d}}$.}
    \textbf{function} disambiguate $(K_d, S_{k_d}, C_a)$:\\
    \ForEach{sense $s_i \in S_{k_d}$}{\label{line:semrelfrom}
        \ForEach{keyword $k_j \in C_a$}{
        $r_j=rel(s_i,k_j)$
        }
        $score_{s_i}=\sum_{j}r_j/|C_a|$
    }\label{line:semrelto}
    $maxScore=max(score_{s_1},\ldots,score_{s_n})$ \label{line:overlapfrom}\\
    \ForEach{sense $s_i \in S_{k_d}$}{
        $newScore=score_{s_i}+(1-maxScore)*overlap(C_a,OC(s_i))$
        $score_{s_i}=newScore$\label{line:overlap}
    }\label{line:overlapto}
    $maxScore=max(score_{s_1},\ldots,score_{s_n})$\label{line:freqfrom} \\
    \ForEach{sense $s_i \in S_{k_d}$}{
        \If{$score_{s_i} > proximityFactor*maxScore$}{
            $newScore=score_{s_i}+(1-maxScore)*normFreq(s_i)$
            $score_{s_i}=newScore$
        }
    }\label{line:freqto}
\end{algorithm}

\begin{enumerate}
    \item \textbf{Applying the semantic relatedness:} First, the algorithm computes an initial disambiguation between the senses in ${S_{k_d}}$ and the active context $C_a$ (Lines~\ref{line:semrelfrom}-\ref{line:semrelto}). For this, we use the updated relatedness measure 
    presented in the previous section (Equations~\ref{eq:termWord} and~\ref{eq:arccos}). The score assigned to each sense (${score_{s_i}}$) is the mean of $rel(s_i,k_j)$ where ${s_i \in S_{k_d}}$ is a candidate sense of the keyword being disambiguated, and ${k_j \in C_a}$ is a keyword in the active context.
    \item \textbf{Calculating the context overlap:} The disambiguation algorithm weights the scores taking into account the overlap between $C_a$ and the 
    ontological context of each sense, ${OC(s_i)}$ (Lines~\ref{line:overlapfrom}-\ref{line:overlapto}). 
    Note that ${OC(s_i)}$ includes its synonyms, glosses, and labels, as well as labels of other related terms, such as hypernyms, hyponyms, meronyms, holonyms, etc. The overlap is calculated (ignoring stopwords) as:
    $$overlap(C_a,OC(s_i)) =\frac{|OC(s_i)\cap C_a|}{min(|{OC(s_i)}|,|C_a|)}$$
    \item \textbf{Frequency of usage:} Finally, 
    the frequency of use of the highest scored senses is taken into account (Lines~\ref{line:freqfrom}-~\ref{line:freqto}), if such information is available. The proximity decision is handled by a ${proximityFactor\in[0,1]}$, which is combined with the maximum of the scores of the senses ($proximityFactor * maxScore$) to obtain a threshold. The scores of the senses $s_i$ which are above such a threshold are then updated using:
    $$normFreq({s_i})=\sqrt{a*\frac{frequency_{s_i}}{\sum_{j}frequency_{s_j}}+b}$$ 
    
    where $\sum_{j}frequency_{s_j}$ is equal to the sum of the frequency of all senses of $k_d$, $a$ and $b$ are constrained\footnote{We set $a = b = 0.5$ and $proximityFactor = 0.75$ as indicated in~\cite{gracia2009multiontology}.} by $a,b \in [0,1]$ and $a + b = 1$.
\end{enumerate}

The output of the disambiguation algorithm is a 
score for each possible sense ${s_i \in S_{k_d}}$ that represents the confidence level of being the right sense according to the active context $C_a$. 
Note that, in our approach, ${S_{k_d}}$ is not restricted to any particular dictionary, as it could be dynamically built from, e.g., different ontological resources.

\subsection{Proposed Modifications}
\label{sec:proposedmodifs}
As our aim is to study the best way to exploit word embeddings, we have analyzed their characteristics and explored different approaches to use them in the adopted disambiguation process. 
In particular, in this section, we present a list of possible modifications to the Step~2 of the algorithm (Lines~\ref{line:overlapfrom}-\ref{line:overlapto}) to include and take advantage of the properties of word embeddings 
along with the rationale behind them. For the rest of the section, let ${maxScore}$ be the maximum score among all senses in ${S_{k_d}}$, ${centroid}$ a function to calculate an average vector by the arithmetic mean of a set of vectors, and ${rel_{w}}$ the angular distance in Equation~\ref{eq:arccos}. 
Thus, the different approaches are described below:

\begin{itemize}
    \item \textbf{Average:} The straightforward way to include the embeddings is to calculate directly the average vector of the different bag of words involved in the disambiguation, under the assumption that the semantically coherent groups of words should outstand from the others. Thus, instead of computing the overlap between the semantic descriptions $OC(s_i)$ of each sense $s_i$ and the current active context $C_a$, we propose to compute the average between the word vectors from ${C_a}$ and ${OC(s_i)}$ to obtain a new score. Line~\ref{line:overlap} in Algorithm~\ref{alg:disambiguation} changes to:
    $$\hspace{0.80cm}newScore(s_i)=score_{s_i}+(1-maxScore)*average({C_a},{OC(s_i)})$$
    where ${average({C_a},{OC(s_i)})}$ is:
    \begin{align*}
    \begin{split}
    average({C_a},{OC(s_i)})=\frac{\sum_{i, j}rel_{w}(k_i,w_j)}{|{C_a}||{OC(s_i)}|}, \\
    (k_i \in {C_a}, w_j \in {OC(s_i)},  i=1..|{C_a}|, j=1..|{OC(s_i)}|)
    \end{split}
    \end{align*}
    That is, we consider each set of words as a cluster in the vector space, and we represent them by their centroid. If there are elements that do not contribute to the semantic cohesion of the clusters, they will contribute negatively (they will increase the semantic distance) to select a particular sense for the target keyword\footnote{We also studied other cluster-based distances measures (e.g., single linkage), but the results did not improve using the centroid-based measure, so we focused on the average vector which is broadly used in the literature.}.
    
    \item \textbf{Sense centroid without most frequent component:} As an evolution of the previous method, we studied the method described by Arora et al.~\cite{arora2016simple}, called \textit{Smooth Inverse Frequency (SIF)}. They propose to represent a sentence by a weighted average vector of its word vectors which the most frequent component using PCA/SVD is substracted from. Thus, we propose to consider the semantic description ${OC(s_i)}$ of all senses of the sense inventory as sentences, and to calculate the SIF embedding of them. Then, during the disambiguation, we compute a new score (Line~\ref{line:overlap}) for the sense being considered ${s_i}$ by measuring the distance between the centroid of the active context ${C_a}$ and the SIF vector of each $OC(s_i)$, following this computation:\vspace{0.10cm}
    
    $\hspace{0.25cm}newScore(s_i)=score_{si}+(1-maxScore)*$
    
    $\vspace{0.15cm}\hspace{2.30cm}rel_{w}(centroid({C_a}), SIF({OC(s_i)}))$
    
    Note that we do not substract the SIF vector from $C_a$ as all its words are already deemed as important. 
    The most frequent component vector we are removing may encompass those words that occur most frequently in a corpus and lack semantic content (e.g., stop-words), 
    thus not contributing to the actual disambiguation.
    
    \item \textbf{Top-K nearest words:} As a variant of the two previous methods, in this method, we select the top \emph{k} nearest words from the semantic description ${OC(s_i)}$ of a sense to ${C_a} \cup k_d$. After that, we compute the distance between centroids of the active context and the top K nearest words selected to obtain its new score: \vspace{0.1cm}
    
    $\hspace{0.25cm}newScore(s_i)=score_{s_i}+(1-maxScore) *$
    
    $\hspace{2.30cm}rel_{w}(centroid({C_a}),centroid($
    
    $\vspace{0.15cm}\hspace{2.8cm}topKNearest(centroid({C_a}),{OC(s_i)})))$
    
    In this case, we work under the same hypothesis as for the selecing an \emph{active context}: the words that belong to the semantic description of the sense that are the closest ones to the active context and the keyword that is being disambiguated, 
    should be the most significant to contribute in making a correct disambiguation. 
    
    \item \textbf{Doc2vec:} Finally, instead of treating the ontological descriptions as bag of words in this method, we consider them as proper documents and apply \textit{doc2vec}~\cite{le2014distributed}. In particular, each semantic description ${OC(s_i)}$ of the senses becomes a document, and \textit{doc2vec} allows to calculate an embedding space for all of them. Then, we compute the distance between the centroid of the active context ${C_a}$ and the embedding calculated for the sense. Note that \textit{doc2vec} learns as well a word embeddings model that it uses during training. We use those word vectors to create the centroid of the active context. Therefore, in a similar way, the new score is computed as:\vspace{0.10cm}
    
    $\hspace{0.25cm}newScore(s_i)=score_{s_i}+(1-maxScore)*$
    
    $\vspace{0.15cm}\hspace{2.30cm}rel_{w}(centroid({C_a}), doc2vec({OC(s_i)}))$
    
    We consider the semantic descriptions as documents 
    to capture the distributional semantics both at local (window) and global (document) scope.
    
\end{itemize}

We report the best results that we obtained by applying these different approaches in the following section.

\section{Experimental Evaluation}
\label{sec:experimentaleval}
In this section, we discuss the results obtained in the experiments that we have carried out to evaluate our proposal. 
Firstly, we evaluated different available embedding models using the distance proposed in Equation~\ref{eq:arccos}. We performed several tests comparing to human judgment in order to check how the \emph{angular distance} behaved. Secondly, we evaluated the potential of our keyword disambiguation algorithm and the relatedness measure among ontology terms and words in the context of WSD, including all algorithm variations that we have proposed in Section~\ref{sec:proposedmodifs}.

For the experiments, we used the following pre-trained vectors: \textit{word2vec} trained on Google News corpus\footnote{\url{https://code.google.com/archive/p/word2vec/}}, 
\textit{word2vec} trained on Wikipedia\footnote{\label{laugithub}\url{https://github.com/jhlau/doc2vec}}, \textit{doc2vec} trained also on Wikipedia\footnote{Dump dated in 2015-12-01.}, \textit{GloVe} trained on Wikipedia 2014 and Gigaword 5 corpus\footnote{\url{https://nlp.stanford.edu/projects/glove/}}, and ${NASARI_{embed}}$ along with the \textit{word2vec} word embeddings trained on the UMBC corpus\footnote{\url{http://lcl.uniroma1.it/nasari/\#two}}. We used WordNet\textsuperscript{\ref{wordnetfootnote}} as the sense inventory.

\subsection{Correlation with Human Judgment}
\label{sec:correlation}
In order to validate the hypothesis of the suitability of using word embeddings along with the use of the \emph{angular distance} to compute semantic relatedness, we first analysed the correlation of such a technique with human judgment in a basic word-to-word comparison. For this purpose, we used 
different datasets available in the literature which contain pairs of words whose relatedness was manually assessed by different people. The datasets and their details can be seen in Table~\ref{tab:datasets}.

\begin{table}[htb!]
    \caption{Correlation with human judgment benchmarks.}
    \centering
    \begin{tabular}{|c|c|c|c|}
    \hline
        Dataset & \#Word Pairs & \#Human Judges \\
        \hline
        MC-30~\cite{miller1991contextual} & 30 & 38 \\
        WordS353-Rel~\cite{finkelstein2001placing} & 252 & 13 \\
        WordS353-Sim~\cite{finkelstein2001placing} & 203  & 16 \\
        RG-65~\cite{rubenstein1965contextual} & 65 & 51 \\
        MEN dataset (train/dev)~\cite{bruni2014multimodal} & 2000/1000 & crowdsourced\footnotemark \\
        GM dataset~\cite{wise08} & 30 & 30 \\
        \hline 
    \end{tabular}
    \label{tab:datasets}
\end{table}
\footnotetext{They used Amazon Mechanical Turk:  \url{https://www.mturk.com/}}

The results obtained for Spearman correlation are presented in Table~\ref{tab:spearmancorrel}. Reported values, where available, were calculated using the widespread used cosine similarity. We can see that using the angular distance (Equation~\ref{eq:arccos}) to calculate relatedness between pairs of words also offers a semantic correlation with the human judgment. 
In particular, regarding the GM dataset~\cite{wise08}, the authors reported a 78\% using the previous \emph{relWeb} measure (Equation~\ref{eq:relweb}). We can see a strong improvement in this dataset by using word embeddings: we achieve up to a 87.3\% using \textit{word2vec} trained on Google News (taking into account the average of all models, we achieve an average of 81.2\% for this dataset). These results enable us to use the angular distance as the core relatedness measure in Equations~\ref{eq:completeeq} to~\ref{eq:level1}. Note that for \textit{word2vec} and \textit{doc2vec} trained on Wikipedia we can not provide a comparison, because Lau \& Baldwin~\cite{lau-baldwin-2016-empirical} did not evaluate the correlation with human judgment.

\begin{table*}[!htb]
  \caption{Spearman correlation coefficients between the angular distance applied on word pairs and human judgment in different datasets. Upper values are our evaluations, lower ones are the reported values in the original papers using the cosine distance. Highlighted values equal or outperform the best result (78\%) for the same dataset in~\cite{wise08}.}
  \label{tab:spearmancorrel}
  \begin{tabular}{|c|c|c|c|c|c|c||c|}
    \hline
    Vectors\textbackslash Datasets & MC-30& WS353-Sim& WS353-Rel& RG-65& MEN& GM & Average\\
    \hline
    GloVe & 70.4& 66.5& 56.1& 76.9& 74.2& \textbf{84.5} & 71.4\\
    \hdashline
    Reported at~Pennington et al.~\cite{pennington2014glove} & 72.7 & 65.8 & - & 77.8 & - & - & 72.1\\
    \hline
    Google News word2vec& 80.0 & 77.2& 63.5& 76.0& 77.0& \textbf{87.3} & 76.8\\
    \hdashline
    Reported at Camacho-Collados et al.~\cite{camacho2016nasari} & 80.0 & 77.0 & - & - & - & - & 78.5\\
    \hline
    ${NASARI_{embed}+UMBCw2v}$& 70.3& 72.7& 56.8& 70.7& 74.5& 74.7 & 70.0\\
    \hdashline
    Reported at Camacho-Collados et al.~\cite{camacho2016nasari} & 83.0 & 68.0 & - & 80.0 & - & - & 75.5\\
    \hline
    Wikipedia word2vec& 80.9& 77.9& 62.2& 78.3& 76.9& \textbf{81.8} & 76.3\\
    \hdashline
    Reported at Lau \& Baldwin~\cite{lau-baldwin-2016-empirical} & - & - & - & - & - & - & -\\
    \hline
    Wikipedia doc2vec& 73.3& 69.0& 52.3& 71.6& 72.0& \textbf{77.8} & 69.3\\
    \hdashline
    Reported at Lau \& Baldwin~\cite{lau-baldwin-2016-empirical} & - & - & - & - & - & - & -\\
    \hline
  \end{tabular}
\end{table*}

\subsection{Word Sense Disambiguation Evaluation}
To evaluate our proposal, we used three datasets oriented to WSD: SemCor 2.0 dataset\footnote{\url{http://web.eecs.umich.edu/\~mihalcea/downloads.html\#semcor}}, SemEval2013 all-words WSD dataset\footnote{\label{semeval2013footnote}\url{https://www.cs.york.ac.uk/semeval-2013/task12.html}}, and SemEval2015 All-Words Sense Disambiguation and Entity Linking dataset\textsuperscript{\ref{semeval2015footnote}}.
We used WordNet\textsuperscript{\ref{wordnetfootnote}} as sense inventory. For SemCor~2.0, we specifically used WordNet~2.0 as such a dataset is annotated with this version, and for the rest of datasets we used WordNet 3.0.

We tested all the options proposed in Section~\ref{sec:proposedmodifs} for the disambiguation algorithm, and we obtained that the \textit{Top-K nearest words} option achieves the best results\footnote{Achieving the best performance for ${K=15}$.}. Thus, due to space restrictions, we focus on \textit{Top-K nearest words} option in this section\footnote{\label{expsurl}The interested reader can find all the details of the experiments at \url{https://bit.ly/2lqCzop}}.  
Regarding the models, we selected \textit{word2vec} trained in Google News and \textit{word2vec} trained in Wikipedia 
because they showed better average correlation with human judgment in different datasets (see Table~\ref{tab:spearmancorrel}); and ${NASARI_{embed}+UMBCw2v}$ word embeddings because, although they do not excelled in correlation with human judgment, they showed the best performance in all test datasets for WSD. Finally, 
in order to compare the results to~\cite{wise08}, we report the precision results for SemCor 2.0; while we report the F-score results for the rest of datasets.

\subsubsection*{SemCor2.0 Experiments:} 
Following~\cite{wise08}, in this set of experiments, for each of three selected highly ambiguous nouns (\textit{plant}, \textit{glass}, and \textit{earth}), we took~10 random sentences from the corpus. Table~\ref{tab:semcor2} presents the results: all cases outperform the results achieved in~\cite{wise08}, which reported an average precision of 57\%. Our best performance is an average precision of 63.15\% with ${NASARI_{embed}+UMBCw2v}$ vectors. 
%
In fact, SIF method shows equal or even slightly better performance in this particular dataset using ${NASARI_{embed}+UMBCw2v}$ vectors and \textit{word2vec} trained in Google News vectors. However, in the rest of cases it is \textit{Top-K nearest words} method that obtains the best results. 
In addition, SIF method requires to preprocess the target sense inventory to calculate the sentence embeddings, introducing a mild dependence to it. 
Our selected method shows a good performance (it improves the results of the original algorithm), while allowing to be more decoupled from the actual sense inventory used.

\begin{table*}[!htb] 
  \caption{Precision results for SemCor 2.0 dataset (10 random sentences) adopting Top-K nearest words. The two rightmost columns show the results using the \textit{relWeb} based relatedness measure and Most Frequent Sense methods as reported in~\cite{wise08}.}
  \label{tab:semcor2}
  \begin{tabular}{|c|c|c|c||c|c|}
    \hline
     Experiment\textbackslash Approach& Wikipedia word2vec & Google News word2vec & ${NASARI_{embed}+UMBCw2v}$ & \textit{relWeb}* & Most Freq. Sense*\\
    \hline
    10 sent. with PLANT& 58.44\%& 63.03\%& 66.20\%& 80\% & 40\%\\
    10 sent. with GLASS& 57.47\%& 63.78\%& 60.15\%& 30\% & 30\%\\
    10 sent. with EARTH& 59.21\%& 56.38\%& 62.33\%& 60\% & 60\%\\
    \hline
    AVERAGE& 58.41\%& 61.13\%& 63.15\%& 57\%& 43\%\\ 
    \hline
  \end{tabular}
\end{table*}

\subsubsection*{SemEval Results:} In Table~\ref{tab:fscore}, we present the results obtained for SemEval 2013 and SemEval 2015. In this case, ${NASARI_{embed}+UMBC}\\{w2v}$ vectors achieved the best results, with an F-score of 64.39\%. In SemEval 2013, UMCC-DLSI reported the best results, with a F-score of 64.7\%, similar to ours. Besides, our results are similar to other state-of-the-art systems using sense embeddings\KILL{vectors}: Camacho et al.~\cite{camacho2016nasari} reported an F-Score value of 66.7\%  with their ${NASARI_{lexical}}$ vectors evaluated in SemEval 2013. Unfortunately, they do not provide results for their ${NASARI_{embed}}$ vectors using WordNet. 
Regarding SemEval 2015, the best reported result in our task reached an F-score of 65.8\%, while the baseline used to compare systems (BabelNet First Sense (BFS)) was an F-score of 67.5\%. We reach an F-score value of 61.61\%, which, while does not beat previous values, is meritory given that our approach is focused on situations where linguistic information might be scarce (e.g., keyword-based input).

\begin{table*}[!htb]
  \caption{F-Score results for SemEval 2013 and 2015 datasets  adopting Top-K nearest words\KILL{and using WordNet 3.0 as sense inventory}. Values in column highlighted by ${\diamond}$ correspond to the best system in each SemEval dataset. Values in column highlighted by ${^\pm}$ are the baselines reported in SemEval 2013~and SemEval 2015.}
  \label{tab:fscore}
  \begin{tabular}{|c|c|c|c||c|c|}
    \hline
     & Wikipedia word2vec & Google News word2vec & ${NASARI_{embed}+UMBCw2v}$ & Best system${^\diamond}$ & Baseline${^\pm}$ \\
    \hline
    SemEval 2013 & 59.59\% & 62.81\% & 64.39\% & 66.7\% & 63.0\%\\
    SemEval 2015 & 61.32\% & 60.37\% & 61.61\% & 65.8\% & 67.5\%\\
    \hline
  \end{tabular}
\end{table*}

To sum up, our proposal improves the results presented in~\cite{wise08} by substituting their Web search engine-based measure to one that uses word embeddings. We also improve the disambiguation results reported in~\cite{gracia2009multiontology} by adapting their algorithm to exploit the properties of the word embeddings. Our proposal achieves similar performance levels to the SOTA, while providing the flexibility to work independently of the resources used (i.e., word embeddings, sense inventory), and reducing the barriers to its application to any domain.

\section{Conclusions and future work}
\label{sec:conclusionsandfuturework}
In this paper, we have presented a keyword disambiguation approach based on a semantic relatedness measure which exploits the semantic information captured by word embeddings and tries to map the meanings from a sense inventory. We have visited the semantic relatedness measure proposed in~\cite{wise08} to adapt it to work with word embeddings instead of relying on Web search engines, and we have improved a disambiguation algorithm~\cite{gracia2009multiontology} by exploring different uses and types of embeddings (both at word and sentence level). 

To validate our proposal, we have performed several experiments around Word Sense Disambiguation (WSD) where we have used different pre-trained word embeddings and WordNet as the resource to obtain the target senses of words. With our proposal:
\begin{itemize}
    \item We are able to relate words and meanings from a sense inventory (e.g., ontology terms) in a flexible way, by exploiting available resources and regardless the domain in which we are working. This makes our measure adaptive and general enough to be used for different contexts. 
    \item We provide a method which can be adapted to any domain in a dictionary-decoupled way, provided that we have a document corpus which would allow us to capture the distributional semantics. This lowers the requirements of data in order to build more specific models for particular domains. 
    \item We have tested the capabilities of different word embedding models, improving the results presented in~\cite{wise08}. We evaluated our measure in the same SemCor 2.0 dataset described in this work, and we have obtained in the best case an average increase of 6\% in precision (a relative improvement of about~11\%).
    \item Being decoupled from a fixed pool of senses does not come at the expense of performance. We achieve similar quality of results than having an ad hoc and more expensive trained model capturing the possible senses. In particular, we have tested our measure in SemEval 2013 and SemEval 2015 datasets reaching an F-score of 64.39\% and 61.61\% respectively. These results are similar to the state of the art~\cite{camacho2016nasari} using \textit{sense2vec} approaches.
\end{itemize}

As future work, we want to further extend our approach to the field of concept discovery (similar to entity search~\cite{Balog:2018:ER}, but focused on concepts rather than on instances). We also want to explore newer contextualized word embeddings, such as BERT or XLNet (based in ELMo~\cite{peters2018deep}), and how they could be used in this context.
Finally, we would like to propose a specific dataset exclusively for keyword disambiguation taking QALD\footnote{QALD is a series of evaluation campaigns on Question Answering over Linked Data: (\url{http://qald.aksw.org}).} datasets as baseline; we want to develop it in order to test our relatedness measure in a more appropriate dataset for the context in which we focus: the disambiguation of keyword-based inputs.


\bibliographystyle{ACM-Reference-Format}
\bibliography{Arxiv-SemRelatedness-biblio} 

\end{document}